\documentclass[pdflatex,sn-mathphys]{sn-jnl}% Math and Physical Sciences Reference Style
\jyear{2022}%

\usepackage{etoolbox}
\makeatletter
\patchcmd{\ps@headings}{Springer Nature 2021 \LaTeX\ template}{}{}{}
\patchcmd{\ps@headings}{Springer Nature 2021 \LaTeX\ template}{}{}{}
\patchcmd{\ps@titlepage}{Springer Nature 2021 \LaTeX\ template}{}{}{}
\makeatother

\theoremstyle{thmstyleone}%
\theoremstyle{thmstyletwo}%

\theoremstyle{thmstylethree}%

\begin{document}

\title[GIDN]{GIDN: A Lightweight Graph Inception Diffusion Network for High-efficient Link Prediction}


\author[1]{\fnm{Zixiao} \sur{Wang}}\email{zwang62@hust.edu.cn}
\author[1]{\fnm{Yuluo} \sur{Guo}}\email{guoyl@hust.edu.cn}
\author[1,2]{\fnm{Jin} \sur{Zhao}}\email{zjin@hust.edu.cn}
\author*[1,2]{\fnm{Yu} \sur{Zhang}}\email{zhyu@hust.edu.cn}
\author[1]{\fnm{Hui} \sur{Yu}}\email{huiy@hust.edu.cn}
\author[1]{\fnm{Xiaofei} \sur{Liao}}\email{xfliao@hust.edu.cn}
\author[2]{\fnm{Biao} \sur{Wang}}\email{wangbiao@zhejianglab.com}
\author[2]{\fnm{Ting} \sur{Yu}}\email{yuting@zhejianglab.com}

\affil[1]{\orgdiv{National Engineering Research Center for Big Data Technology and System, Services Computing Technology and System Lab, Cluster and Grid Computing Lab, School of Computer Science and Technology}, \orgname{Huazhong University of Science and Technology}, \orgaddress{\city{Wuhan},\country{China}}}

\affil[2]{\orgname{Zhejiang Lab}, \orgaddress{\city{Hangzhou}, \country{China}}}


\abstract{In this paper, we propose a Graph Inception Diffusion Networks(GIDN) model. This model generalizes graph diffusion in different feature spaces, and uses the inception module to avoid the large amount of computations caused by complex network structures. We evaluate GIDN model on Open Graph Benchmark(OGB) datasets, reached an 11\% higher performance than AGDN on ogbl-collab dataset.}

\maketitle
\section{Introduction}\label{sec1}

Currently, many implementations of semantic applications rely on knowledge graphs, where new knowledge is inferred from existing knowledge. e.g., recommendation computing~\cite{SMR,yu2017knowledge} and search engines~\cite{bhagdev2008hybrid,arenas2016faceted}. Knowledge graphs map real-world entities and relationships between entities into triples~\cite{hogan2021knowledge}, denoted as (head entity, relation, tail entity), e.g., (Turing, BornIn, England), indicating that Turing was born in England. Knowledge graphs have gained great achievement in many fields, such as natural language processing~\cite{al2020named,annervaz2018learning}, deep learning~\cite{xie2016representation}, and graph processing~\cite{zheng2018question,zhang2018variational}. 

\par Even though large-scale Knowledge Graphs currently contain more than a billion triples, these datasets are still incomplete~\cite{ji2021survey}, meaning that they are missing a large number of valid triples. Knowledge graph embedding, also known as link prediction, can effectively complement these datasets~\cite{nickel2015review,wang2019explainable}. For example, it is known that cephalosporins and alcohol together can cause a disulfiram-like reaction, and cefpirome is a cephalosporin, so it is inferred that cefpirome and alcohol cannot be taken together. Knowledge graph embedding aims to encode real-world entities and relationships as a low-dimensional vector that can be efficiently stored and computed, and contain semantic-level information~\cite{chami2020low}.

\par Due to the rapid development of artificial intelligence,  knowledge graphs that meet the need for data processing and understanding are receiving increasing attention. The main approach of some well-designed knowledge graph models is to map entities and relations into and reason in a continuous space. For example, TransE~\cite{bordes2013translating} interprets relations as translations of entities in low dimensions, modeling the inversion and composition patterns. RotatE~\cite{sun2019rotate} treats relations as rotations from source entities to target entities in complex spatial vectors, enabling modeling and inference of various relational patterns. There are also methods in symbolic space where models are based on symbolic logic rules, such as Markov logic networks~\cite{richardson2006markov}. Recently, researchers have proposed to combine neural methods and symbolic methods to make models with better capacity and interpretability, such as pLogicNet~\cite{qu2019probabilistic} and NBFNet~\cite{zhu2021neural}.

\section{Realted Work}\label{sec2}
\subsection{Graph Neural Networks}
 To address the drawbacks of heuristic techniques, graph neural networks provide heuristics that can learn appropriate features from the networks themselves in the link prediction task~\cite{bruna2013spectral,niepert2016learning,dai2016discriminative}. SEAL~\cite{zhang2018link} fixes multiple flaws of WLNM, not only replacing the fully connected neural network in WLNM with a graph neural network, but also allowing learning from latent and explicit node features, thereby absorbing multiple types of information. Because some information is lost when computing feature vectors at the pooling layer, LGLP~\cite{cai2021line} proposes to learn features directly from target links instead of closed subgraphs. And in order for the graph convolutional layer to effectively learn edge embeddings from the graph, LGLP modifies the original bounding subgraph into the corresponding line graph. SEAL needs to operate on each subgraph, which doesn't work very well on large graphs. NBFNet~\cite{zhu2021neural} explicitly captures the path between two nodes for link prediction while achieving relatively low time complexity. 
\subsection{Shallow Embedding Technique}
Inspired by the word2vec~\cite{church2017word2vec}, Perozzi et al.~\cite{perozzi2014deepwalk} proposes a graph mining approach combining random walk~\cite{tong2006fast} and word2vec algorithm to learn the hidden information and social representation of a graph. Tang et al.~\cite{tang2015line} puts forward an algorithm whose target function preserves both local and global structure of a graph, resulting in greater efficiency and effectiveness. Grover et al.~\cite{grover2016node2vec} improves the random walk method in Deepwalk~\cite{perozzi2014deepwalk}, makes it possible to reflect the characteristics of both depth-first and breadth-first sampling, thus improving the effectiveness of network embedding. However, one limitation of these works is that too many parameters are associated in the model, which significantly reduces the computational speed.
\subsection{Logic Rule Induction Methods}
The heuristic technology is calculated according to the characteristics of the graph structure, and the key is to calculate the similarity calculation score for the neighborhood of the two target nodes~\cite{nowell2003link,lu2011link}. As shown in Figure 1, according to the maximum number of neighbor hops used in the calculation process, the heuristic methods can be divided into three groups, including first-order, second-order and higher-order heuristics. Zhang et al. ~\cite{zhang2017weisfeiler} proposed Weisfeiler-Lehman Neural Machine(WLNM) to extract a subgraph for each target link domain as an adjacency matrix, and train a fully connected neural network on these adjacency matrices to learn a link prediction model.  Studies have shown that higher-order heuristics such as rooted PageRank~\cite{brin2012reprint} and Simrank~\cite{jeh2002simrank} have better performance than lower-order heuristics~\cite{adamic2003friends,zhou2009predicting}. However, the increase in the number of grid hops means higher computational costs and memory consumption. Moreover, the lack of heuristics has poor applicability to different types of networks~\cite{lu2011link, klein1993resistance}, which means complex computation is required to find the appropriate heuristic for different networks.

\section{Method}\label{sec3}

% \subsection{Graph Diffusion Networks}
Encapsulating the representation of graph diffusion can provide a better basis for prediction than the graph itself. Graph diffusion is the use of a matrix for each prediction target in a graph to represent the information in its proximity. For example, for node n, its H-hop node is the node reached by jumping H times from node n. The i-th row of the matrix is the neighbourhood information of the i-hop of node n. Standard graph diffusion operations rely on extensive tensor calculations, which require expensive storage space and running time. Graph Diffusion Networks model uses a combination of small-hop nodes and learnable generalised weighting coefficients to achieve multi-layer generalised graph diffusion in different feature spaces, while also ensuring moderate complexity and running time.

% \subsection{Inception Module}
If the depth of the network is increased, the network will become computationally complex. The Inception module is able to capture rich features while avoiding the computational effort associated with an overly deep network, making it more adaptable to training with a large number of samples.

% \subsection{Random Walk Augmentation}
Data augmentation can expand the dimension of training data. In graph structure, data augmentation mainly focus on nodes and edges. Random walk is a method removing the edges between nodes who have different labels and building connections for the nodes with the same labels.

\section{Evaluation}\label{sec4}
We conduct experiments on the ogbl-collab dataset~\cite{hu2020open}. We run the model 10 times and record the mean ± standard deviation of Hits@50. The results are shown in Table~\ref{tab1}. Our method reaches better performance than AGDN~\cite{sun2020adaptive}, and higher ceiling than PLNLP~\cite{wang2021pairwise}.

\begin{table}[h]
\begin{center}
\begin{minipage}{174pt}
\caption{Performance of GIDN}\label{tab1}%
\begin{tabular}{c|c}
\toprule
Method & ogbl-collab(Hits@50)  \\
\midrule
AGDN    & 0.4480 ± 0.0542  \\
PLNLP   & 0.7059 ± 0.0029  \\
GIDN    & 0.7096 ± 0.0055  \\
\botrule
\end{tabular}
\end{minipage}
\end{center}
\end{table}

 \section{Acknowledgments}
 This work is supported by National Natural Science Foundation of China under grant No. 62072193, and Major Scientific Research Project of Zhejiang Lab No. 2022PI0AC03.


\begin{thebibliography}{10}
\providecommand{\url}[1]{#1}
\csname url@samestyle\endcsname
\providecommand{\newblock}{\relax}
\providecommand{\bibinfo}[2]{#2}
\providecommand{\BIBentrySTDinterwordspacing}{\spaceskip=0pt\relax}
\providecommand{\BIBentryALTinterwordstretchfactor}{4}
\providecommand{\BIBentryALTinterwordspacing}{\spaceskip=\fontdimen2\font plus
\BIBentryALTinterwordstretchfactor\fontdimen3\font minus
  \fontdimen4\font\relax}
\providecommand{\BIBforeignlanguage}[2]{{%
\expandafter\ifx\csname l@#1\endcsname\relax
\typeout{** WARNING: IEEEtranS.bst: No hyphenation pattern has been}%
\typeout{** loaded for the language `#1'. Using the pattern for}%
\typeout{** the default language instead.}%
\else
\language=\csname l@#1\endcsname
\fi
#2}}
\providecommand{\BIBdecl}{\relax}
\BIBdecl

\bibitem{SMR}Gong, F., Wang, M., Wang, H., Wang, S. \& Liu, M. SMR: medical knowledge graph embedding for safe medicine recommendation. {\em Big Data Research}. \textbf{23} pp. 100174 (2021)
\bibitem{yu2017knowledge}Yu, T., Li, J., Yu, Q., Tian, Y., Shun, X., Xu, L., Zhu, L. \& Gao, H. Knowledge graph for TCM health preservation: Design, construction, and applications. {\em Artificial Intelligence In Medicine}. \textbf{77} pp. 48-52 (2017)
\bibitem{bhagdev2008hybrid}Bhagdev, R., Chapman, S., Ciravegna, F., Lanfranchi, V. \& Petrelli, D. Hybrid search: Effectively combining keywords and semantic searches. {\em European Semantic Web Conference}. pp. 554-568 (2008)
\bibitem{arenas2016faceted}Arenas, M., Grau, B., Kharlamov, E., Marciuška, Š. \& Zheleznyakov, D. Faceted search over RDF-based knowledge graphs. {\em Journal Of Web Semantics}. \textbf{37} pp. 55-74 (2016)
\bibitem{hogan2021knowledge}Hogan, A., Blomqvist, E., Cochez, M., D’Amato, C., Melo, G., Gutierrez, C., Kirrane, S., Gayo, J., Navigli, R., Neumaier, S. \& Others Knowledge graphs. {\em ACM Computing Surveys (CSUR)}. \textbf{54}, 1-37 (2021)
\bibitem{al2020named}Al-Moslmi, T., Ocaña, M., Opdahl, A. \& Veres, C. Named entity extraction for knowledge graphs: A literature overview. {\em IEEE Access}. \textbf{8} pp. 32862-32881 (2020)
\bibitem{annervaz2018learning}Annervaz, K., Chowdhury, S. \& Dukkipati, A. Learning beyond datasets: Knowledge graph augmented neural networks for natural language processing. {\em ArXiv Preprint ArXiv:1802.05930}. (2018)
\bibitem{ji2021survey}Ji, S., Pan, S., Cambria, E., Marttinen, P. \& Philip, S. A survey on knowledge graphs: Representation, acquisition, and applications. {\em IEEE Transactions On Neural Networks And Learning Systems}. \textbf{33}, 494-514 (2021)
\bibitem{zheng2018question}Zheng, W., Yu, J., Zou, L. \& Cheng, H. Question answering over knowledge graphs: question understanding via template decomposition. {\em Proceedings Of The VLDB Endowment}. \textbf{11}, 1373-1386 (2018)
\bibitem{zhang2018variational}Zhang, Y., Dai, H., Kozareva, Z., Smola, A. \& Song, L. Variational reasoning for question answering with knowledge graph. {\em Thirty-second AAAI Conference On Artificial Intelligence}. (2018)
\bibitem{wang2019explainable}Wang, X., Wang, D., Xu, C., He, X., Cao, Y. \& Chua, T. Explainable reasoning over knowledge graphs for recommendation. {\em Proceedings Of The AAAI Conference On Artificial Intelligence}. \textbf{33}, 5329-5336 (2019)
\bibitem{xie2016representation}Xie, R., Liu, Z., Jia, J., Luan, H. \& Sun, M. Representation learning of knowledge graphs with entity descriptions. {\em Proceedings Of The AAAI Conference On Artificial Intelligence}. \textbf{30} (2016)
\bibitem{nickel2015review}Nickel, M., Murphy, K., Tresp, V. \& Gabrilovich, E. A review of relational machine learning for knowledge graphs. {\em Proceedings Of The IEEE}. \textbf{104}, 11-33 (2015)
\bibitem{chami2020low}Chami, I., Wolf, A., Juan, D., Sala, F., Ravi, S. \& Ré, C. Low-dimensional hyperbolic knowledge graph embeddings. {\em ArXiv Preprint ArXiv:2005.00545}. (2020)
\bibitem{bordes2013translating}Bordes, A., Usunier, N., Garcia-Duran, A., Weston, J. \& Yakhnenko, O. Translating embeddings for modeling multi-relational data. {\em Advances In Neural Information Processing Systems}. \textbf{26} (2013)
\bibitem{sun2019rotate}Sun, Z., Deng, Z., Nie, J. \& Tang, J. Rotate: Knowledge graph embedding by relational rotation in complex space. {\em ArXiv Preprint ArXiv:1902.10197}. (2019)
\bibitem{richardson2006markov}Richardson, M. \& Domingos, P. Markov logic networks. {\em Machine Learning}. \textbf{62}, 107-136 (2006)
\bibitem{qu2019probabilistic}Qu, M. \& Tang, J. Probabilistic logic neural networks for reasoning. {\em Advances In Neural Information Processing Systems}. \textbf{32} (2019)
\bibitem{zhu2021neural}Zhu, Z., Zhang, Z., Xhonneux, L. \& Tang, J. Neural bellman-ford networks: A general graph neural network framework for link prediction. {\em Advances In Neural Information Processing Systems}. \textbf{34} pp. 29476-29490 (2021)
\bibitem{church2017word2vec}Church, K. Word2Vec. {\em Natural Language Engineering}. \textbf{23}, 155-162 (2017)
\bibitem{perozzi2014deepwalk}Perozzi, B., Al-Rfou, R. \& Skiena, S. Deepwalk: Online learning of social representations. {\em Proceedings Of The 20th ACM SIGKDD International Conference On Knowledge Discovery And Data Mining}. pp. 701-710 (2014)
\bibitem{tong2006fast}Tong, H., Faloutsos, C. \& Pan, J. Fast random walk with restart and its applications. {\em Sixth International Conference On Data Mining (ICDM'06)}. pp. 613-622 (2006)
\bibitem{tang2015line}Tang, J., Qu, M., Wang, M., Zhang, M., Yan, J. \& Mei, Q. Line: Large-scale information network embedding. {\em Proceedings Of The 24th International Conference On World Wide Web}. pp. 1067-1077 (2015)
\bibitem{grover2016node2vec}Grover, A. \& Leskovec, J. node2vec: Scalable feature learning for networks. {\em Proceedings Of The 22nd ACM SIGKDD International Conference On Knowledge Discovery And Data Mining}. pp. 855-864 (2016)
\bibitem{zhang2017weisfeiler}Zhang, M. \& Chen, Y. Weisfeiler-lehman neural machine for link prediction. {\em Proceedings Of The 23rd ACM SIGKDD International Conference On Knowledge Discovery And Data Mining}. pp. 575-583 (2017)
\bibitem{zhang2018link}Zhang, M. \& Chen, Y. Link prediction based on graph neural networks. {\em Advances In Neural Information Processing Systems}. \textbf{31} (2018)
\bibitem{adamic2003friends}Adamic, L. \& Adar, E. Friends and neighbors on the web. {\em Social Networks}. \textbf{25}, 211-230 (2003)
\bibitem{klein1993resistance}Klein, D. \& Randić, M. Resistance distance. {\em Journal Of Mathematical Chemistry}. \textbf{12}, 81-95 (1993)
\bibitem{nowell2003link}Nowell, D. The link prediction problem for social networks. {\em Proc Of 12th International Conference On Information And Knowledge Management, 2003}. (2003)
\bibitem{lu2011link}Lü, L. \& Zhou, T. Link prediction in complex networks: A survey. {\em Physica A: Statistical Mechanics And Its Applications}. \textbf{390}, 1150-1170 (2011)
\bibitem{brin2012reprint}Brin, S. \& Page, L. Reprint of: The anatomy of a large-scale hypertextual web search engine. {\em Computer Networks}. \textbf{56}, 3825-3833 (2012)
\bibitem{jeh2002simrank}Jeh, G. \& Widom, J. Simrank: a measure of structural-context similarity. {\em Proceedings Of The Eighth ACM SIGKDD International Conference On Knowledge Discovery And Data Mining}. pp. 538-543 (2002)
\bibitem{zhou2009predicting}Zhou, T., Lü, L. \& Zhang, Y. Predicting missing links via local information. {\em The European Physical Journal B}. \textbf{71}, 623-630 (2009)
\bibitem{cai2021line}Cai, L., Li, J., Wang, J. \& Ji, S. Line graph neural networks for link prediction. {\em IEEE Transactions On Pattern Analysis And Machine Intelligence}. (2021)
\bibitem{bruna2013spectral}Bruna, J., Zaremba, W., Szlam, A. \& LeCun, Y. Spectral networks and locally connected networks on graphs. {\em ArXiv Preprint ArXiv:1312.6203}. (2013)
\bibitem{niepert2016learning}Niepert, M., Ahmed, M. \& Kutzkov, K. Learning convolutional neural networks for graphs. {\em International Conference On Machine Learning}. pp. 2014-2023 (2016)
\bibitem{dai2016discriminative}Dai, H., Dai, B. \& Song, L. Discriminative embeddings of latent variable models for structured data. {\em International Conference On Machine Learning}. pp. 2702-2711 (2016)

\bibitem{sun2020adaptive}Sun, C. \& Wu, G. Adaptive graph diffusion networks with hop-wise attention. {\em ArXiv Preprint ArXiv:2012.15024}. (2020)
\bibitem{hu2020open}Hu, W., Fey, M., Zitnik, M., Dong, Y., Ren, H., Liu, B., Catasta, M. \& Leskovec, J. Open graph benchmark: Datasets for machine learning on graphs. {\em Advances In Neural Information Processing Systems}. \textbf{33} pp. 22118-22133 (2020)
\bibitem{wang2021pairwise}Wang, Z., Zhou, Y., Hong, L., Zou, Y. \& Su, H. Pairwise Learning for Neural Link Prediction. {\em ArXiv Preprint ArXiv:2112.02936}. (2021)

\end{thebibliography}
\end{document}